\documentclass[conference]{IEEEtran}
\IEEEoverridecommandlockouts

\usepackage{amssymb}
\usepackage{amsmath}
\usepackage{amsthm}
\usepackage{caption}
\usepackage{derivative}
\usepackage{multirow}
\usepackage{booktabs}
\usepackage{comment}
\usepackage{lineno}
\usepackage{adjustbox}
\usepackage{standalone}
\usepackage{float}
\usepackage{cite}
\usepackage{amsmath,amssymb,amsfonts}
\usepackage{algorithmic}
\usepackage{graphicx}
\usepackage{textcomp}
\usepackage{xcolor}

\def\BibTeX{{\rm B\kern-.05em{\sc i\kern-.025em b}\kern-.08em
T\kern-.1667em\lower.7ex\hbox{E}\kern-.125emX}}

\renewcommand\thesubsubsection{\Alph{subsection}.\arabic{subsubsection}}

\begin{document}

\title{Any-to-Any Vision-Language Model for Multimodal X-ray Imaging and Radiological Report Generation}
\author{
    \IEEEauthorblockN{Daniele Molino\IEEEauthorrefmark{1},
                      Francesco di Feola\IEEEauthorrefmark{2},
                      Linlin Shen\IEEEauthorrefmark{3},
                      Paolo Soda\IEEEauthorrefmark{1}\IEEEauthorrefmark{2},
                      and
                      Valerio Guarrasi\IEEEauthorrefmark{1}}
    \IEEEauthorblockA{\IEEEauthorrefmark{1}Research Unit of Computer Systems and Bioinformatics, Department of Engineering, \\
    Università Campus Bio-Medico di Roma, Rome, Italy \\
    Email: daniele.molino@unicampus.it, p.soda@unicampus.it, valerio.guarrasi@unicampus.it}
    \IEEEauthorblockA{\IEEEauthorrefmark{2}Department of Diagnostics and Intervention, Biomedical Engineering and Radiation Physics, Umeå University, Umeå, Sweden \\
    Email: francesco.feola@umu.se, paolo.soda@umu.se}
    \IEEEauthorblockA{\IEEEauthorrefmark{3}College of Computer Science and Software Engineering, Shenzhen University, Shenzhen, China \\
    Email: llshen@szu.edu.cn}
}


\maketitle
\begin{abstract}
Generative models have revolutionized Artificial Intelligence (AI), particularly in multimodal applications. 
However, adapting these models to the medical domain poses unique challenges due to the complexity of medical data and the stringent need for clinical accuracy.
In this work, we introduce a framework specifically designed for multimodal medical data generation.
By enabling the generation of multi-view chest X-rays and their associated clinical report, it bridges the gap between general-purpose vision-language models and the specialized requirements of healthcare.
Leveraging the MIMIC-CXR dataset, the proposed framework shows superior performance in generating high-fidelity images and semantically coherent reports.
Our quantitative evaluation reveals significant results in terms of FID and BLEU scores, showcasing the quality of the generated data. 
Notably, our framework achieves comparable or even superior performance compared to real data on downstream disease classification tasks, underlining its potential as a tool for medical research and diagnostics.
This study highlights the importance of domain-specific adaptations in enhancing the relevance and utility of generative models for clinical applications, paving the way for future advancements in synthetic multimodal medical data generation.
\end{abstract}


\section{Introduction}
\label{Introduction}
Artificial Intelligence (AI) is transforming various domains, including healthcare, by enabling the analysis of complex medical data and facilitating advancements such as precise diagnostics and personalized treatments~\cite{alowais2023revolutionizing}. 
Among these, multimodal data integration represents a significant breakthrough, allowing the synthesis of complementary information from diverse sources for enhanced diagnostic insights~\cite{di2025graph,guarrasi2024systematic,guarrasi2024multimodal,guarrasi2023multi}.
Despite this potential, AI adoption in healthcare is hindered by critical challenges, related to data scarcity and privacy constraints~\cite{alzubaidi2023survey}. 
Training deep learning (DL) models typically demands large, high-quality data, but the existing medical datasets are often limited in size and diversity. 
This insufficiency not only impacts model performance, causing bias or overfitting, but also restricts the generalization capabilities of these models to unseen clinical scenarios. 
Privacy regulations, such as the GDPR in Europe~\cite{gdpr2016general} and HIPAA in the United States~\cite{act1996health}, add another layer of complexity by limiting data sharing and collaboration.
To mitigate these issues, synthetic data generation has emerged as a promising solution, enabling the creation of artificial datasets that replicate the complexity and variability of real-world medical data. By bypassing constraints related to data availability and privacy, these techniques offer a practical solution for training robust AI models in healthcare.
Generative AI has advanced significantly since the introduction of Generative Adversarial Networks (GANs)~\cite{goodfellow2020generative}, which pioneered data synthesis through adversarial learning.
While GANs achieved early success, their limitations, including training instability and mode collapse, have posed challenges, particularly in domains requiring high precision, such as medical imaging~\cite{rofena2024deep}.
Recently, Diffusion Models (DMs)~\cite{ho2020denoising} have emerged as a more reliable approach, leveraging a stepwise denoising framework to produce diverse, high-fidelity data.
Build on DM success, Latent Diffusion Models (LDMs)~\cite{rombach2022high} extend this capability by operating within a compressed latent space, reducing computational demands and enabling advanced control over the generation process.
By employing encoders to condition the generation on relevant features, such as anatomical structures or disease markers, LDMs facilitate the tailored synthesis of data for various diagnostic and research needs~\cite{ramesh2022hierarchical, saharia2022photorealistic}.
However, despite significant advancements brought by LDMs, research in multimodal generation remains unexplored, particularly in addressing the challenges of integrating diverse modalities into a cohesive framework. 
In this context, Composable Diffusion (CoDi)~\cite{tang2023anytoany} stands out for its innovative approach to multimodal data generation. 
CoDi’s unified latent space enables consistent and coherent any-to-any generation across modalities, avoiding the limitations posed by sequential methods, which achieve multimodal generation through a multi-step approach. 
However, while CoDi excels in general-purpose applications, its adaptation to the medical domain remains underexplored.
In this paper, we introduce $\text{CoDi}_{\mathit{{XR}}}$, an adaptation of CoDi designed to address the specific challenges of the medical field. 
By tailoring CoDi’s framework to the nuances of medical data, $\text{CoDi}_{\mathit{{XR}}}$ aims to bridge the gap between general-purpose multimodal generation and the specialized requirements of healthcare applications. 
Our work highlights the importance of domain-specific adaptation in enhancing the relevance and utility of generative models for medical use cases.
Specifically, we evaluate $\text{CoDi}_{\mathit{{XR}}}$’s performance on the MIMIC-CXR dataset, focusing on its ability to generate multi-view Chest X-ray (CXR) images and their associated clinical reports. 
This comprehensive evaluation underscores the model’s potential to address key challenges in medical data generation, including consistency across X-rays views and the alignment between visual and textual modalities.

\section{Materials}

\begin{table}[t]
\caption{Pathology label distribution for the X-rays classification task.}
\label{tab:PosDist}
\centering
\resizebox{1\columnwidth}{!}{
\begin{tabular}{cccc}
\toprule
\textbf{Pathology} & \textbf{\# of Samples} & \textbf{\# of Negatives} & \textbf{\# of Positives} \\ \midrule
Atl. & 10561 & 9123 & 1438 \\
Cgml. & 10305 & 9159 & 1146 \\
Cnsl. & 9911 & 9657 & 254 \\
Edm. & 10230 & 9456 & 774 \\
Enl. & 9215 & 9132 & 83 \\
Les. & 9476 & 9134 & 342 \\
Opc. & 11136 & 9156 & 1980 \\
Eff. & 10558 & 9300 & 1258 \\
Pnm. & 10454 & 9601 & 853 \\
Ptx. & 9337 & 9253 & 84 \\
\bottomrule
\end{tabular}
}
\end{table}

To achieve our purpose, it is crucial to leverage a multimodal medical dataset that captures the complementary nature of different modalities, such as imaging and textual data. 
Such datasets are essential for training models capable of synthesizing clinically accurate and coherent outputs across diverse medical data types.
We used the MIMIC-CXR~\cite{johnson2019mimic} dataset that contains 377.110 CXR images along with their corresponding radiology reports, for a total of 227.827 studies conducted at the Beth Israel Deaconess Medical Center in Boston, MA, USA.
Images are acquired in frontal and lateral projection, offering distinct yet complementary diagnostic information~\cite{santosh2018angular}. 
For example, cardiovascular structures and the diaphragm can obscure up to 15\% of the lung, making certain pathologies undetectable in the frontal view alone~\cite{raoof2012interpretation}. 
The lateral view, providing a different perspective, enables the visualization of lesions or abnormalities hidden behind these anatomical structures, thus ensuring more accurate diagnosis~\cite{hashir2020quantifying}.
For these reasons, we treated frontal and lateral CXRs as distinct modalities.
Each radiology report in the dataset is divided in two sections: a finding section that provides a detailed description of both normal and abnormal features observed in the corresponding CXR, and an impression section, which provides a concise summary of the findings intended to support medical decision-making.
In this work, we focused exclusively on the latter, as it offers a concise yet powerful summary of the patient’s condition and it also complies with our text encoder, which, following implementation \cite{tang2023anytoany}, poses a limitation on the length of the report to 77 tokens.
\begin{figure}[t]
\centering
\includegraphics[width=1\columnwidth]{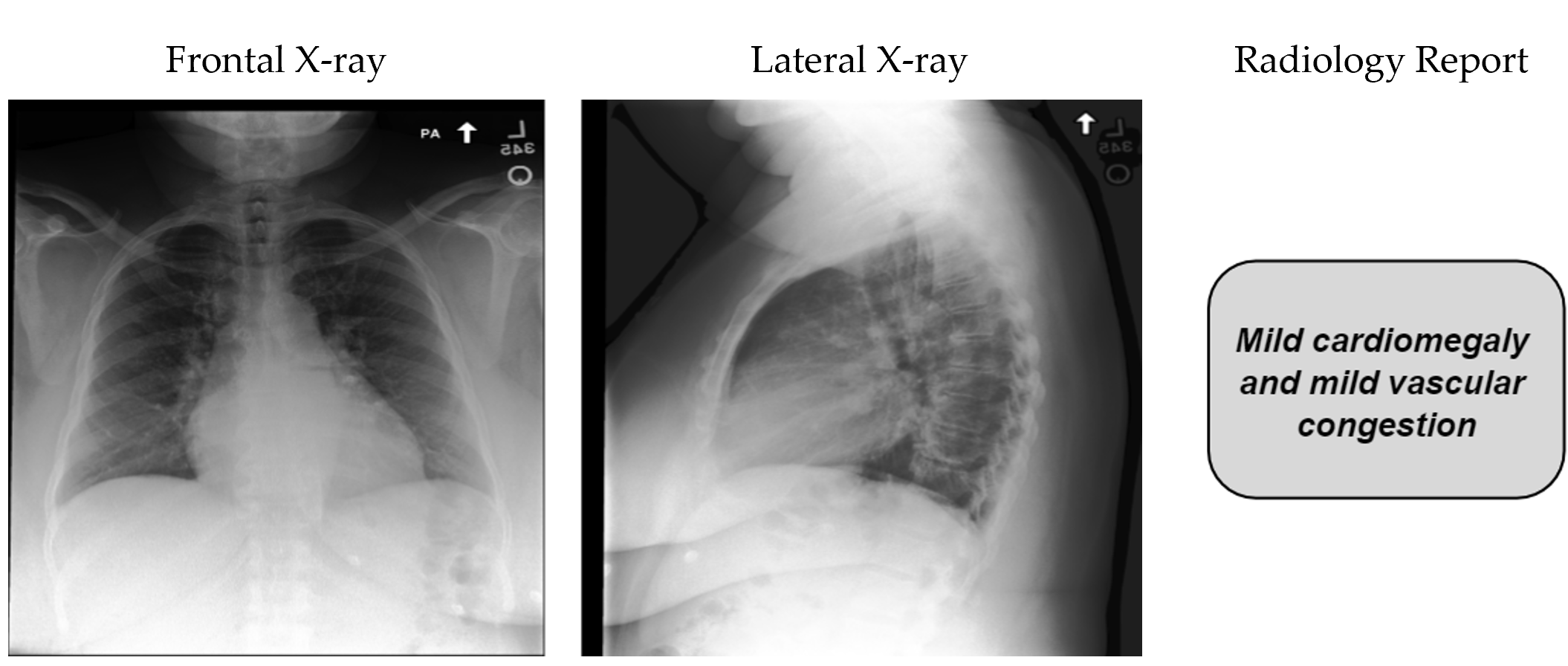}
\caption{A sample of a frontal X-ray, a lateral X-ray and the corresponding radiology report.}\label{fig:mimic}
\end{figure}
Furthermore, we used the original uncompressed X-rays stored in DICOM format~\cite{bidgood1997understanding} to ensure unintended loss of information due to compression. The X-ray's preprocessing involved several steps to standardize and prepare the data for model training.
First, we examinated the pixel spacing of each image and resampled those with non-standard spacing to $[0.139, 0.139]$.
For images with a Photometric Intepretation of Monochrome-1, the pixel values were inverted to ensure proper representation. 
Subsequently, we normalized the images by dividing every pixel by the maximum pixel value possible given by their bit depth, bringing the range to $[0, 1]$. 
Since the original scans are not square, we added zero-padding around the images and then resized them to $256\times256$, to standardize the input size while preserving the integrity of the visual content.
From the repository, we extracted a total of 154.721 X-rays from 78.584 studies, including all the patients for which the radiology report and both frontal and lateral view were present.
To ensure an unbiased evaluation, we extracted an holdout test set before any training procedure. 
This set consists of 33.588 samples, that were carefully selected to guarantee no patient's overlapping between training and test set.
In addition to imaging and textual data, MIMIC-CXR includes diagnostic labels for a set of predefined pathologies, automatically extracted using a rule-based labeler\cite{irvin2019chexpert}. 
In order to comply with recent state-of-the-art practices, we adopted the classifier from \cite{cohen2022torchxrayvision} for evaluation. 
Since this model is trained exclusively on a specific subset of pathologies, we restricted our evaluation to the same subset to ensure a fair and consistent comparison.
Specifically this subset is composed of: Atelectasis (Atl.), Cardiomegaly (Cmgl.), Consolidation (Cnsl.), Edema (Edm.), Enlarged Cardiomediastinum (Enl.), Lung Lesion (Les.), Lung Opacity (Opc.), Pleural Effusion (Eff.), Pneumonia (Pnm.), and Pneumothorax (Ptx.). 
\tablename~\ref{tab:PosDist} reports the number of samples for every pathology in our test set.
An explicative example of a triplet is depicted in Fig.~\ref{fig:mimic}.

\section{Methods}
\label{Methods}
In this work, we focus on adapting CoDi's framework to the medical domain by enabling the generation of multi-view chest X-rays and the corresponding clinical reports.
Let ${{X}}=\{{F}, {L}\}$ represent the two different X-rays modalities, and let ${{R}}$ denote the radiological report modality.
\begin{figure*}[h]
\centering
\includegraphics[width=1\textwidth]{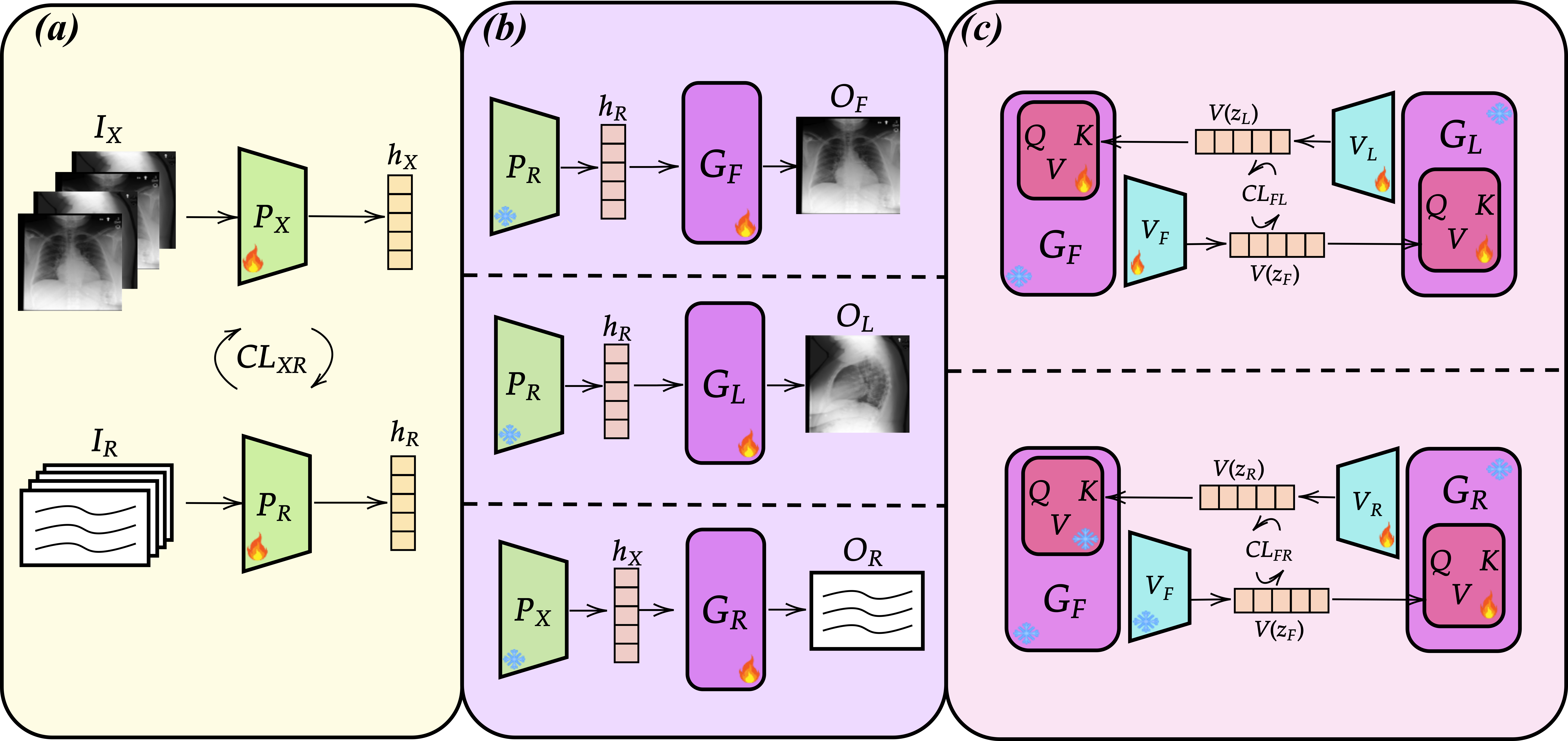}
\captionsetup{justification=raggedright, singlelinecheck=false}
\caption{\textbf{Framework of CoDi}$_{\text{\textbf{\textit{{XR}}}}}$: a) Shared Latent Space construction: Input modalities are processed by modality-specific encoders to extract feature representations, which are aligned in a shared latent space. b) Single modality generation: An LDM for each modality is trained to generate synthetic data from the latent representations extracted by the Prompt Encoders. c) Cross-modal alignment: it ensures consistency and alignment across jointly generated output modalities. Elements marked with a fire icon receive parameter updates during the specific training phase, while those marked with an ice icon remain frozen.}\label{fig:model}
\end{figure*}
The architecture of $\text{CoDi}_{\mathit{{XR}}}$, illustrated in Fig.\ref{fig:model}, consists of three main components: in panel (a), the input modalities $I$ are processed by prompt encoders $P$ to extract feature representations, which are aligned in a shared latent space using contrastive learning. 
This alignment ensures that the latent representations capture the underlying correlations between frontal and lateral X-rays and radiological reports.
In panel (b), we perform single modality generation by training modality-specific LDMs $G$ conditioned on the aligned latent representations. 
This training allows the model to synthesize outputs $O$ that reflect the features and anomalies observed in the inputs.
In panel (c), cross-modal alignment is performed to enhance consistency and coherence between generated outputs.
The following sections provide a detailed explanation of each component.

\subsection{Building a Shared Latent Space}
\label{Prompt Encoder Training}
We propose to align any input modalities within a shared latent space by leveraging contrastive learning. 
This approach allows the model to be freely conditioned on any input combination, even those absent in the training data.
Inspired by~\cite{tang2023anytoany}, we take advantage of an efficient technique called Bridging Alignment to align the representations extracted by modality-specific prompt encoders. 
Given that our three modalities consist of texts and images, we adopt the Contrastive Language-Image Pretraining (CLIP)~\cite{radford2021learning} approach, which leverages an image encoder and a text encoder, denoted as $P_{X}$ and $P_{R}$, jointly trained on large-scale datasets of text-image pairs. 
Their representations are then aligned through contrastive learning, using the InfoNCE contrastive loss~\cite{oord2018representation}:
\begin{equation}
\label{eq:infonce}
    {{\cal L}_{A,B} =  - \log \frac{{\exp ({h_A^i}^{\top}{h_B^i}/\tau )}}{{\exp ({h_A^i}^{\top}{h_B^i}/\tau ) + \sum\nolimits_{j \ne i} {\exp ({h_A^i}^{\top}{h_B^i}/\tau )}}}}
\end{equation}
where $A,B$ denotes two generic modalities, $\tau$ is the scalar temperature regulating the softness of the softmax distribution, and $i$,$j$ refers to positive and negative pairs, respectively. 
By doing so, the encoders learn a shared representation space that effectively captures the semantics of both modalities.
In order to reduce the computational overload, we used a single a ViT-base\cite{dosovitskiy2020image} encoder for both frontal and lateral X-rays feature representations, while we leverage a masked self-attention Transformer \cite{tang2023anytoany} for the text encoding.
Following Fig~\ref{fig:model}.a, let ${I_X}$ be a batch of both frontal and lateral X-rays and let ${I_R}$ be their corresponding reports, we obtain the embeddings $h_{X}=P_{X}({I_X})$, $h_{R}=P_{R}({I_R})$ for both modalities, aligned through the symmetric contrastive loss $\mathit{CL}_{\mathit{{XR}}}={\mathcal{L}_{X, R}}+{\mathcal{L}_{R, X}}$, which aim to make the embeddings close together in the latent space.

\subsection{Single modality generation through Latent Diffusion Model}
\label{Multi-Prompt Training}
Diffusion models \cite{ho2020denoising} are a class of generative models, inspired by non-equilibrium thermodynamics, designed to learn a distribution by simulating the diffusion of information over time. 
Let $x$ be a data point, during the training procedure, random noise is progressively added to $x$, while the model learns to reverse this process. 
During inference, the model can generate new data points by denoising samples drawn from simple distributions.
LDMs~\cite{rombach2022high}, an extension of this framework, operates on a latent space representation $z$ of the data $x$, significantly reducing the computational overhead, as they work in a lower-dimensional space.
In the LDM architecture, an autoencoder is initially trained to reconstruct the input data $x$, i.e., $\hat{x} = D(E(x))$, where $E$ and $D$ represent the encoder and decoder, respectively.
The latent variable $z = E(x)$ is then progressively diffused across time steps $t$ according to a predefined variance schedule $\beta_1,\dots, \beta_T$, i.e., $q(z_t|z_{t-1}) = \mathcal{N}(z_t; \sqrt{1-\beta_t} z_{t-1}, \beta_t I)$ \cite{ dickstein2015deep,ho2020denoising}.
The forward process enables the random sampling of $z_t$ at any timestep in a closed form \cite{dickstein2015deep,ho2020denoising}: $z_t = \alpha_t z + \sigma_t \epsilon$, where $\epsilon \sim \mathcal{N}(0, I)$, $\alpha_t := 1-\beta_t$ and $\sigma_t := 1 - \prod_{s=1}^t\alpha_s$. 
Following the reparametrization method proposed in \cite{ho2020denoising}, the denoising training objective can be expressed as \cite{rombach2022high}:
\begin{equation}
\label{eq:df}
    \mathcal{L}_{D} = \mathbb{E}_{z, \epsilon, t}\|\epsilon - \epsilon_{\theta}(z_t, t, P(I))\|_2^2
\end{equation}
where $t \sim \mathcal{U}[1, T]$ denotes the diffusion time step sampled from a uniform distribution, $\epsilon_{\theta}$ is a denoising model with a UNet architecture parameterized by $\theta$, $I$ represents the conditional variable used to control the generation process and $P$ denotes the prompt encoder.
The conditioning mechanism operates by first converting $I$ into a feature representation $P(I)$, which is then used to condition the UNet $\epsilon_{\theta}$ through cross-attention, as described in \cite{rombach2022high}. 
Developing a generative model capable of handling multiple inputs and outputs simultaneously requires extensive and diverse training data while maintaining high-quality outputs across all modalities. 
To tackle these complexities, $\text{CoDi}_{\mathit{{XR}}}$ employs a modular and flexible design, allowing individual LDMs to be trained independently for specific modalities before seamlessly integrating them into a cohesive system.
The LDMs for image generation, $G_{F}$ and $G_{L}$, adopts the same architecture of Stable Diffusion 1.5~\cite{rombach2022high}, where AutoKL~\cite{esser2021taming} is used as the variational autoencoder (VAE) to map the input modalities in a lower dimensional space. 
As stated in \cite{chambon2022adapting}, the most effective approach for the adaptation of an LDM to the medical imaging domain is to fine-tune the UNet component. 
Therefore, we kept the VAE frozen and only trained the UNet.
We trained the two LDMs independently, using a batch size of 512, a learning rate of \(5 \times 10^{-5}\), and a weight decay of \(1 \times 10^{-4}\). 
Both models were trained for 100 epochs using the AdamW optimizer.
For text generation, the UNet architecture of $G_\mathbf{R}$ is based on~\cite{xu2023versatile}, which introduced the fully-connected residual blocks (FCResBlock). 
These expand the 768-dimensional text latent vectors into a 320-by-4 hidden feature and follow the residual block paradigm with GroupNorms~\cite{wu2018group}, SiLU~\cite{elfwing2018sigmoid}, and skip connections. 
We adopt Optimus~\cite{li2020optimus} as the text VAE, which consist of a BERT~\cite{devlin2018bert} text encoder and a GPT-2~\cite{radford2019language} text decoder. 
Unlike the LDMs used for X-ray images, we decided to fine-tune both the VAE and the UNet in two separate training rounds. 
This approach is necessary as the model has to effectively adapt to a completely different vocabulary.
Following~\cite{li2020optimus}, the training process begins with a reconstruction task, where the VAE is tasked with accurately reconstructing input radiology reports from their latent representations.
Once the first step is fulfilled, the UNet is trained for report generation using a batch size of 1024 and a learning rate equal to \(1 \times 10^{-5}\). 
The weight decay, the optimizer configuration and the number of epochs remained consistent with those used for the X-ray LDMs.

\subsection{Multi-output generation via Cross-modal Latent Alignement}
\label{Multi-Output Generation}
The third training stage enables the simultaneous generation of any combination of output modalities, ensuring that each generative flow is aware of the others.
To this end, we incorporate two trainable components into each LDM:
the first is an encoder $V$, that projects the latent variable of the diffusion process $z$ into a shared latent space; the second is a cross-attention layer $QKV$, that allows each LDM to attend to the generative process of another model.
Formally, let us consider frontal and lateral X-rays being jointly synthesized by $G_{{F}}$ and $G_{L}$ and let $z_{F}$ and $z_{L}$ denote their latent variables at a generic diffusion step, respectively.
Following Fig~\ref{fig:model}.c, the encoder $V_{L}$ first projects $z_{L}$ into a shared latent space. 
Then, in each layer of $G_{F}$, the cross-attention layer attends to $V_{L}(z_{L})$. 
For the diffusion model of modality ${F}$, the training objective in Eq.\ref{eq:df} becomes: 
\begin{equation}
\mathcal{L}_{D}^{{F}} = \mathbb{E}_{z, \epsilon, t}\|\epsilon - \epsilon_{\theta_c}(z_{F}, V_{L}(z_{L}), t, P(I_{R}))\|_2^2,
\end{equation}
where $\theta_c$ represents the parameters of the cross-attention layer in the UNet.
The diffusion training objective for the joint generation of ${F}$ and ${L}$ becomes $\mathcal{L}_{D}^{{F}}+{L}_{D}^{{L}}$, where ${L}_{D}^{{L}}$ is defined as:
\begin{equation}
\mathcal{L}_{D}^{{L}} = \mathbb{E}_{z, \epsilon, t}\|\epsilon - \epsilon_{\theta_c}(z_{L}, V_{F}(z_{F}), t, P(I_{R}))\|_2^2.
\end{equation}
To further enhance the training process, we incorporated contrastive learning between the outputs of the environmental encoders, using again the InfoNCE contrastive
loss, as defined in Eq.\ref{eq:infonce}.
The total loss for the alignment of $G_{F}$ and $G_{L}$ is $\mathcal{L}_{\mathit{FL}}=\mathcal{L}_{D}^{{F}}+\mathcal{L}_{D}^{L}+\mathit{CL}_{\mathit{FL}}$.
By training only these two additional components while keeping the rest of the model frozen, $\text{CoDi}_{\mathit{{XR}}}$ effectively learns to generate multiple modalities simultaneously while maintaining high-quality outputs.
As it is possible to see at the bottom of Fig~\ref{fig:model}.c, once this training phase is completed, an analogous process is applied for the simultaneous generation of \(O_F\) and \(O_R\). 
However, at this stage, the entire \(G_{F}\) module, including the $QKV$ and the $V_{F}$ encoder, can be kept frozen, allowing the model to focus exclusively on aligning textual outputs with the visual features extracted from the X-rays, where the training objective is $\mathcal{L}_{\mathit{FR}}=\mathcal{L}_{D}^{{F}}+\mathcal{L}_{D}^{{R}}+\mathit{CL}_{\mathit{FR}}$, where $\mathcal{L}_{D}^{{R}}$ is defined as:
\begin{equation}
\mathcal{L}_{D}^{{R}} = \mathbb{E}_{z, \epsilon, t}\|\epsilon - \epsilon_{\theta_c}(z_{R}, V_{F}(z_{F}), t, P(I_L))\|_2^2.
\end{equation}

\section{Experimental Configuration}
\subsection{Computational Analysis} 
We provide a detailed breakdown of the number of parameters for each model component, to quantify the computational cost of our framework.
The CLIP model, used to extract the latent representations from the input modalities, contains 737 million parameters. 
AutoKL, used as the VAE for the X-ray LDMs, has 101 million parameters, with two instances used in our framework, while Optimus, used as the VAE for the report LDM, consists of 241 million parameters. 
The UNet of the X-ray LDM has 1.77 billion parameters, with two instances used while the Report UNet model has 2.04 billion parameters. 
In total, the number of parameters for all components amounts to 6.77 billion.
All experiments were conducted on a high-performance computing cluster equipped with four NVIDIA A100 GPUs. 

\subsection{Evaluation Metrics}
\label{Metrics}
We conducted both quantitative and qualitative assessments to evaluate the performance of our approach. 
The first focuses on the statistical properties of the generated data, while the second ensures that the outputs accurately align with the expected clinical informations.

\subsubsection{Quantitative Assessment}
To objectively evaluate the quality of the generated outputs, we employed two well-established quantitative metrics, the FID score and the BLEU score, that measure the statistical similarity between synthetic and real data, as well as the linguistic coherence of generated clinical reports. 

The Fréchet Inception Distance (FID)~\cite{heusel2017gans} measures the dissimilarity between real and synthetic samples in the feature space of an Inception-V3 model pre-trained on ImageNet~\cite{russakovsky2015imagenet}, ranging in the interval $[0, +\infty)$, with lower values indicating greater similarity.
However, because the Inception-Net is not trained on a medical dataset, it may lead to misleading results~\cite{tronchin2021evaluating}.
To address this limitation, we computed the FID also with another backbone, i.e., XRV-DenseNet~\cite{cohen2022torchxrayvision}, an in-domain classification model trained to detect pathologies in CXR scans.

The Bilingual Evaluation Understudy (BLEU) compares machine-generated text to a set of references by calculating the n-gram overlap between the two~\cite{papineni2002bleu}, ranging in the interval $[0, +\infty)$.
Following the literature, here we computed the BLEU score for a number of n-grams equal to {1,2,3,4}.
BLEU-1 and BLEU-2 place greater emphasis on the consistency of the vocabulary used, focusing on single words or word pairs, while BLEU-3 and BLEU-4 provide information about the semantic structure of the reports.

\subsubsection{Factual Correctness Assessment}
Ensuring the factual correctness of the generated data is a crucial aspect of evaluating the performance of $\text{CoDi}_{\mathit{{XR}}}$.  
By using well-established classification models and rule-based tools, we assess how well the synthetic outputs align with real-world diagnostic information.

To evaluate whether the models are capable of generating images that accurately reflect the information of the corresponding clinical reports, we classified the generated samples using the XRV-DenseNet~\cite{cohen2022torchxrayvision}, a well-established classifier for CXR classification.
Since this classifier is trained only on a subset of pathologies, we computed the AUC and F1 score for the diseases reported in \tablename~\ref{tab:PosDist}, along with micro, macro, and weighted averages.
The micro average aggregates contributions from all classes to provide an overall measure, the macro average computes the metric independently for each class and averages them, and the weighted average adjusts the macro average by accounting for the number of samples per class.
However, because not all scans have a defined label for every pathology, we computed the performance for each class only when a ground truth was available.

For report classification, we leveraged the CheXpert-Labeler~\cite{irvin2019chexpert}, a rule-based natural language processing tool that reads a text report and extracts whether it mentions the presence or absence of significant radiologic findings.
Since a rule-based classifier is used, it is not possible to compute the AUC; instead, we reported the F1 score for the same subset of disease previously introduced along with No Finding (No F.) class.
To remain consistent with the previous setup, we also reported the micro, macro, and weighted averages for the F1 score.
This task quantifies the ability of the model to generate reports that align with the medical conditions seen in the X-ray images, ensuring that the synthetic reports accurately reflect the diagnostic information provided by the images.

\begin{table*}[t]
\caption[FID Score for X-rays Generation.]{FID score for X-ray generation, with lower values indicating greater similarity. XRV and v3 refers to the two backbones used to compute the score, respectively
XRV-Densenet and Inception-v3.}
\label{tab:FID}
\centering
\resizebox{\textwidth}{!}{
\begin{tabular}{lcccccccccccc}
\toprule
\multirow{2}{*}{\textbf{Model}} & \multicolumn{2}{c}{\textbf{T$\rightarrow$F}} & \multicolumn{2}{c}{\textbf{L$\rightarrow$F}} & \multicolumn{2}{c}{\textbf{L$+$T$\rightarrow$F}} & \multicolumn{2}{c}{\textbf{T$\rightarrow$L}} & \multicolumn{2}{c}{\textbf{F$\rightarrow$L}} & \multicolumn{2}{c}{\textbf{F$+$T$\rightarrow$L}} \\
\cmidrule(lr){2-3} \cmidrule(lr){4-5} \cmidrule(lr){6-7} \cmidrule(lr){8-9} \cmidrule(lr){10-11} \cmidrule(lr){12-13}
& \textbf{v3} & \textbf{XRV} & \textbf{v3} & \textbf{XRV} & \textbf{v3} & \textbf{XRV} & \textbf{v3} & \textbf{XRV} & \textbf{v3} & \textbf{XRV} & \textbf{v3} & \textbf{XRV} \\
\midrule
\multirow{2}{*}{\begin{tabular}[c]{@{}l@{}} CoDi \\ $\text{CoDi}_{\mathit{{XR}}}$ \end{tabular}}       
& 541.44 & 107.23 & 520.03 & 83.12 & 539.35 & 107.17 & 522.00 & 83.01 & 540.65 & 105.63 & 525.80 & 82.52 \\
& \textbf{10.56} & \textbf{0.86} & \textbf{34.89} & \textbf{3.31} & \textbf{22.63} & \textbf{1.89} & \textbf{13.90} & \textbf{0.84} & \textbf{43.24} & \textbf{4.95} & \textbf{23.12} & \textbf{1.99} \\
\bottomrule
\end{tabular}
}
\end{table*}

\begin{table*}[h]
\caption{AUROC and F1-score for CXR classification performance with XRV Densenet.}
\label{tab:CXR_classification}
\centering
\begin{adjustbox}{width=1\textwidth}
\addtolength{\tabcolsep}{-4pt}  
{\scriptsize 
\begin{tabular}{c|cccccccccc|ccc|cccccccccc|ccc}
\toprule
\multirow{2}{*}{\textbf{Model}} & \multicolumn{13}{c|}{\textbf{AUROC} $\uparrow$} & \multicolumn{13}{c}{\textbf{F1-Score} $\uparrow$} \\
\cmidrule(lr){2-14} \cmidrule(lr){15-27}
&\textbf{Atel.} & \textbf{Cmgl.} & \textbf{Cnsl.} & \textbf{Edm.} & \textbf{Enl.} & \textbf{Les.} & \textbf{Opac.} & \textbf{Eff.} & \textbf{Pnm.} & \textbf{Ptx.} & \textbf{Micro} & \textbf{Macro} & \textbf{Weighted} 
& \textbf{Atl.} & \textbf{Cmgl.} & \textbf{Cnsl.} & \textbf{Edm.} & \textbf{Enl.} & \textbf{Les.} & \textbf{Opc.} & \textbf{Eff.} & \textbf{Pnm.} & \textbf{Ptx.} & \textbf{Micro} & \textbf{Macro} & \textbf{Weighted} \\ 
\midrule
Real Data & .84 & .91 & .91 & .93 & .81 & .78 & .85 & .95 & .78 & .86 & .87 & .86 & .87 
& .49 & .49 & \textbf{.35} & .60 & .09 & \textbf{.12} & .60 & .73 & .35 & \textbf{.21} & .49 & .40 & .53 \\
CoDi & .50 & .50 & .50 & .50 & .50 & .50 & .50 & .50 & .50 & .50 & .50 & .54 & .50 
& .0 & .0 & .0 & .0 & .0 & .0 & .0 & .0 & .0 & .0 & .0 & .0 & .0 \\
$\text{CoDi}_{\mathit{{XR}}}$ T$\rightarrow$F & \textbf{.91} & \textbf{.95} & \textbf{.93} & \textbf{.96} & \textbf{.84} & \textbf{.82} & \textbf{.91} & \textbf{.97} & \textbf{.84} & \textbf{.88} & \textbf{.90} & \textbf{.90} & \textbf{.91} 
& \textbf{.60} & \textbf{.60} & .33 & \textbf{.66} & \textbf{.10} & .05 & \textbf{.64} & \textbf{.78} & \textbf{.38} & .17 & \textbf{.53} & .43 & \textbf{.59} \\
$\text{CoDi}_{\mathit{{XR}}}$ L$\rightarrow$F & .84 & .88 & .90 & .91 & .81 & .76 & .85 & .95 & .76 & .82 & .83 & .85 & .90 
& .51 & .45 & .32 & .42 & .06 & .07 & .59 & .73 & .27 & .10 & .43 & .35 & .49 \\
$\text{CoDi}_{\mathit{{XR}}}$ T$+$L$\rightarrow$F & .89 & .91 & .92 & .91 & .82 & .80 & .86 & \textbf{.97} & .77 & .85 & .88 & .89 & .90 
& .54 & .55 & .32 & .64 & .09 & .06 & .63 & .75 & \textbf{.38} & .17 & \textbf{.53} & \textbf{.44} & .58 \\
\bottomrule
\end{tabular}
}
\end{adjustbox}
\end{table*}

\begin{table*}[h]
\caption{BLEU scores for report generation, with higher values indicating greater similarity.}
\label{tab:BLEU}
\centering
\begin{adjustbox}{width=1\textwidth}
\addtolength{\tabcolsep}{-4pt} 
{\scriptsize
\begin{tabular}{c|cccc|cccc|cccc}
\toprule
\multirow{2}{*}{\textbf{Model}} & \multicolumn{4}{c|}{\textbf{F$\rightarrow$T}} & \multicolumn{4}{c|}{\textbf{L$\rightarrow$T}} & \multicolumn{4}{c}{\textbf{F$+$L$\rightarrow$T}} \\ 
\cmidrule(lr){2-5} \cmidrule(lr){6-9} \cmidrule(lr){10-13} 
& \textbf{BLEU-1} & \textbf{BLEU-2} & \textbf{BLEU-3} & \textbf{BLEU-4} 
& \textbf{BLEU-1} & \textbf{BLEU-2} & \textbf{BLEU-3} & \textbf{BLEU-4} 
& \textbf{BLEU-1} & \textbf{BLEU-2} & \textbf{BLEU-3} & \textbf{BLEU-4} \\
\midrule
CoDi
& 0.02 & 0.001 & 0.00 & 0.00     
& 0.02 & 0.01 & 0.00 & 0.00 
& 0.02 & 0.01 & 0.00 & 0.00 \\
$\text{CoDi}_{\mathit{{XR}}}$
& \textbf{0.38} & \textbf{0.27} & \textbf{0.22} & \textbf{0.18}     
& \textbf{0.38} & \textbf{0.28} & \textbf{0.23} & \textbf{0.18} 
& \textbf{0.42} & \textbf{0.32} & \textbf{0.27} & \textbf{0.22} \\
\bottomrule
\end{tabular}}
\end{adjustbox}
\end{table*}

\begin{table*}[h]
\caption{F1-score for report classification performance with CheXpert-Labeler.}
\label{tab: Report-Clf}
\begin{adjustbox}{width=1\textwidth}
\addtolength{\tabcolsep}{-4pt}  
\begin{tabular}{c|ccccccccccc|ccc}
\toprule
\textbf{F1-Score} $\uparrow$ & \textbf{Atl.} & \textbf{Cmgl.} & \textbf{Cnsl.} & \textbf{Edm.}  & \textbf{Enl.} & \textbf{Les.} & \textbf{Opc.} & \textbf{Eff.} & \textbf{Pnm.} & \textbf{Ptx.} & \textbf{No F.} & \textbf{Micro} & \textbf{Macro} & \textbf{Weighted} \\ \midrule
CoDi & .0 & .0 & .0 & .0 & .0 & .0 & .0 & .0 & .0 & .0 & .0 & .0 & .0 & .0 \\
$\text{CoDi}_{\mathit{{XR}}}$ F$\rightarrow$T & .56 & .56 & .15 & .67 & \textbf{.10} & .26 & .47 & .71 & .52 & \textbf{.37} & .88 & .67 & .44 & .67 \\
$\text{CoDi}_{\mathit{{XR}}}$ L$\rightarrow$T & .54 & .58 & .14 & .65 & .07 & .23 & .45 & .75 & .51 & .30 & .89 & .68 & .41 & .68 \\
$\text{CoDi}_{\mathit{{XR}}}$ F$+$L$\rightarrow$T & \textbf{.61} & \textbf{.61} & \textbf{.17} & \textbf{.70} & .09 & \textbf{.31} & \textbf{.53} & \textbf{.78} & \textbf{.59} & .27 & \textbf{.91} & \textbf{.72} & \textbf{.45} & \textbf{.72} \\
\bottomrule
\end{tabular}
\end{adjustbox}
\end{table*}

\section{Results}
\label{sec:results}
In this section, we present the results of $\text{CoDi}_{\mathit{{XR}}}$'s performance evaluation, focusing our analysis on the generation of multi-view chest X-rays and associated radiology reports, comparing $\text{CoDi}_{\mathit{{XR}}}$ against its baseline competitor, CoDi. 

\subsection{Chest X-Ray Generation}
\begin{figure}[t]
\centering
\includegraphics[width=1\columnwidth]{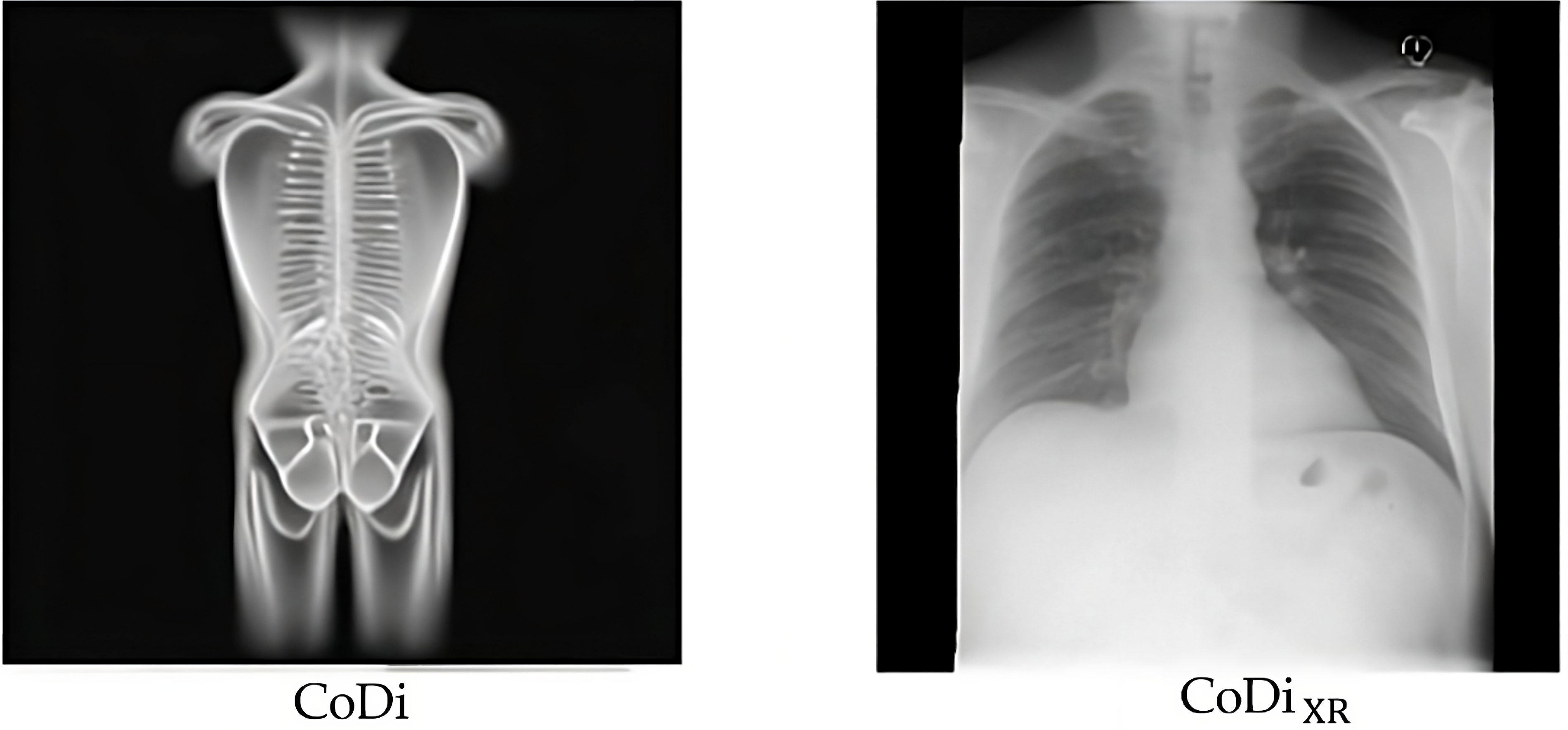}
\caption{Generation comparison between CoDi and $\text{CoDi}_{\mathit{{XR}}}$ using the same textual prompt, i.e., ``X-ray presents no acute cardiopulmonary process''.}\label{fig:comparison}
\end{figure}

Table~\ref{tab:FID} shows the FID scores on the test set for the generation of frontal (F) and lateral (L) X-rays, evaluated using both Inception-V3 and XRV-DenseNet backbones.
Different generation setting have been evaluated: from clinical report (T) to frontal or lateral CXR (T→F, T→L), from lateral or frontal CXR to the other view (L→F, F→L) and from a combination of a clinical report and a CXR image to the other view (L+T→F, F+T→L).
$\text{CoDi}_{\mathit{{XR}}}$ consistently outperforms CoDi across all modalities and generation settings, with significant reductions in FID scores, indicating higher similarity between generated and real images.
It is worth noting that the excessively high FID values for CoDi highlight the crucial role of finetuning the LDMs.
This observation is further supported by Figure \ref{fig:comparison}, which shows one example of generation by CoDi and $\text{CoDi}_{\mathit{{XR}}}$ with the same textual prompt, where the former fails to generate any resemblance of an X-ray.
However, it is possible to observe that $\text{CoDi}_{\mathit{{XR}}}$'s performance tends to degrade slightly when receiving an image as input. 
For instance, in the case of T$\rightarrow$F, the model achieves an excellent FID score of 0.86, but this increases to 3.31 in the L$\rightarrow$F setting. Although the FID values remain low overall, this indicates a slight decline in the quality of the generated X-rays. 
This behavior can be attributed to the training procedure, as the diffusion model is solely trained to generate X-rays from textual input. 
As a consequence, while the model demonstrates impressive zero-shot capabilities in cross-view translation tasks like L$\rightarrow$F, a minor degradation in performance is still noticeable.
Table~\ref{tab:CXR_classification} presents the AUROC and F1 scores for the classification of generated X-rays across 10 pathologies. $\text{CoDi}_{\mathit{{XR}}}$ significantly outperforms CoDi in all cases.
These results indicate that $\text{CoDi}_{\mathit{{XR}}}$ generated images are not only visually realistic but also diagnostically informative.
CoDi exhibits AUROC scores of 0.50 across all pathologies, indicating that the generated X-rays lack any diagnostic value.
This result highlights the limit of CoDi to generate clinically meaningful images. 
Furthermore, as observed in Table~\ref{tab:FID}, $\text{CoDi}_{\mathit{{XR}}}$'s performance slightly degrades when conditioned only on an X-ray. 
For instance, the AUROC score for L$\rightarrow$F generation is lower compared to T$\rightarrow$F, suggesting a minor decline in diagnostic quality.
Moreover, in some cases, $\text{CoDi}_{\mathit{{XR}}}$ outperforms real data. 
We attribute this phenomenon to the model's strong understanding of disease-specific features, which makes the generated samples more easily classifiable by the pre-trained classifier.
However, this result should not be interpreted as entirely positive: it may also suggest that the model has primarily learned to reproduce only the most prototypical or overt manifestations of a pathology, while still struggling to capture more subtle or ambiguous cases that are harder to identify, even for human experts.

\subsection{Report Generation}
Table~\ref{tab:BLEU} presents the BLEU score on the test set for report generation across three different generation settings: frontal CXR to report (F→T), lateral CXR to report (L→T) and both frontal and lateral CXR to report (F+L→T).
These scores highlight the ability of each method to generate reports in comparison to the reference ones, with higher BLEU scores indicating better performances.
The results show that $\text{CoDi}_{\mathit{{XR}}}$ outperform CoDi across all BLEU score metrics.
Notably, CoDi exhibits extremely poor performance, with BLEU scores close to zero, further emphasizing its inability to generate meaningful or coherent reports.
In addition, results suggest that $\text{CoDi}_{\mathit{{XR}}}$ not only utilizes the same terminology as real clinical reports (BLEU-1 and BLEU-2), but it also generates sentences whose structure is consistent with that of actual reports (BLEU-3 and BLEU-4). 
Table~\ref{tab: Report-Clf} reports the F1-scores for disease classification based on the generated reports. 
$\text{CoDi}_{\mathit{{XR}}}$ demonstrates superior alignment between generated reports and the clinical conditions described in the input images. 
The results highlight $\text{CoDi}_{\mathit{{XR}}}$'s remarkable capability to generate clinically meaningful reports. 
This indicates that the report generated by $\text{CoDi}_{\mathit{{XR}}}$ are not only linguistically coherent but also diagnostically reliable, capturing essential clinical details.
Moreover, the results underscore the impact of conditioning on both frontal and lateral X-rays (F$+$L$\rightarrow$T).
In this setting, $\text{CoDi}_{\mathit{{XR}}}$ achieves the highest micro, macro and weighted F1 scores, further emphasizing the importance of leveraging complementary information from multiple views. 
The combination of views enhances the model's ability to detect subtle pathologies and provide more comprehensive diagnostic insights.
In contrast, CoDi fails entirely in this task, with F1 scores equal to 0.0 across all categories, reflecting its inability to generate reports that align with real-world clinical conditions. 
This difference illustrates the effectiveness of domain-specific adaptations in $\text{CoDi}_{\mathit{{XR}}}$, which enable it to bridge the gap between visual data and textual descriptions effectively.

\section{Conclusions}
This work demonstrates the critical role of domain-specific adaptations in tailoring generative models for medical applications. 
By adapting the CoDi framework to the medical domain, $\text{CoDi}_{\mathit{{XR}}}$ effectively leverages multimodal data to generate high-fidelity chest X-rays and clinically coherent reports.
The results underscore the potential of combining LDMs with contrastive learning for addressing the unique challenges posed by medical data, such as ensuring consistency across modalities and preserving clinical relevance.
Despite these advancements, certain limitations highlight the need for further development. 
Notably, $\text{CoDi}_{\mathit{{XR}}}$ exhibits a performance drop when conditioned solely on X-rays, indicating that the current training procedure, focused primarily on text-to-image generation, requires improvement to fully exploit multimodal conditioning. 
Addressing this limitation could unlock even greater potential for cross-view and cross-modal tasks~\cite{ruffini2024multi}.
Future research will focus on enhancing the training framework to improve performance in all generation settings while exploring broader applications in diverse medical imaging and diagnostic scenarios. 
Additionally, we aim to conduct a comprehensive evaluation involving clinical experts, in order to assess the quality and clinical plausibility of the generated samples from a domain-specific, human-centric perspective.
Moreover, we plan to benchmark $\text{CoDi}_{\mathit{XR}}$ against state-of-the-art unimodal models on specific generation tasks (e.g., text-to-image or image-to-text), to better understand the benefits of adopting a unified any-to-any multimodal approach over specialized unimodal pipelines. 
These developments could establish $\text{CoDi}_{\mathit{{XR}}}$ as a foundational tool for synthetic data generation in the healthcare domain.

\section*{Acknowledgments}
Daniele Molino is a Ph.D. student enrolled in the National Ph.D. in Artificial Intelligence, XL cycle, course on Health and life sciences, organized by Università Campus Bio-Medico di Roma.
This work was partially founded by: 
i) Università Campus Bio-Medico di Roma under the program ``University Strategic Projects'' within the project ``AI-powered Digital Twin for next-generation lung cancEr cAre (IDEA)''; 
ii) PNRR MUR project PE0000013-FAIR.
iii)  Cancerforskningsfonden Norrland project MP23-1122;
iv) Kempe Foundation project JCSMK24-0094; 
v) the Italian Ministry of Foreign Affairs and International Cooperation, grant number PGR01156
Resources are provided by the National Academic Infrastructure for Supercomputing in Sweden (NAISS) and the Swedish National Infrastructure for Computing (SNIC) at Alvis @ C3SE, partially funded by the Swedish Research Council through grant agreements no. 2022-06725 and no. 2018-05973.

\bibliographystyle{unsrt}
\bibliography{biblio.bib}

\end{document}